\documentclass[letterpaper]{article} %
\usepackage{aaai25}  %
\usepackage{times}  %
\usepackage{helvet}  %
\usepackage{courier}  %
\usepackage[hyphens]{url}  %
\usepackage{graphicx} %
\usepackage{natbib}  %
\usepackage{caption} %
\usepackage{algorithm}
\usepackage{algorithmic}

\usepackage{newfloat}
\usepackage{listings}
\DeclareCaptionStyle{ruled}{labelfont=normalfont,labelsep=colon,strut=off} %
\floatstyle{ruled}
\newfloat{listing}{tb}{lst}{}
\floatname{listing}{Listing}
\usepackage{amsmath, amsfonts, amssymb, amsthm}
\usepackage{physics}
\newtheorem{definition}{Definition}
\newtheorem*{example}{Example}

\usepackage{adjustbox}
\usepackage{booktabs}

\usepackage{tikz}
\usetikzlibrary{shapes.geometric, positioning, backgrounds, fit, arrows.meta}

\title{Rashomon in the Streets: Explanation Ambiguity in Scene Understanding}
\author {
    Helge Spieker\textsuperscript{\rm 1},
    Jørn Eirik Betten\textsuperscript{\rm 1},
    Arnaud Gotlieb\textsuperscript{\rm 1},
    Nadjib Lazaar\textsuperscript{\rm 2},
    Nassim Belmecheri\textsuperscript{\rm 1}  
}
\affiliations {
\textsuperscript{\rm 1}Simula Research Laboratory, Oslo, Norway\\
\textsuperscript{\rm 2}LISN, Université Paris-Saclay, Saclay, France\\
\{helge,jorneirik,nassim,arnaud\}@simula.no, lazaar@lisn.fr\\
}

\begin{document}

\maketitle

\begin{abstract}
Explainable AI (XAI) is essential for validating and trusting models in safety-critical applications like autonomous driving.
However, the reliability of XAI is challenged by the Rashomon effect, where multiple, equally accurate models can offer divergent explanations for the same prediction. 
This paper provides the first empirical quantification of this effect for the task of action prediction in real-world driving scenes. 
Using Qualitative Explainable Graphs (QXGs) as a symbolic scene representation, we train Rashomon sets of two distinct model classes: interpretable, pair-based gradient boosting models and complex, graph-based Graph Neural Networks (GNNs). 
Using feature attribution methods, we measure the agreement of explanations both within and between these classes. Our results reveal significant explanation disagreement. 
Our findings suggest that explanation ambiguity is an inherent property of the problem, not just a modeling artifact. 
\end{abstract}

\section{Introduction}

Scene understanding in automated driving is the task of perceiving the vehicle’s environment, including other traffic participants like cyclists and pedestrians, and predicting their actions.
This capability is critical for monitoring interactions between traffic participants, anticipating their future behavior, and making informed, safe decisions \cite{yang_scene_2019,muhammad_deep_2021,muhammad_vision-based_2022}.

Automated driving is a challenging context and its user acceptance requires sufficient trust into the reliability of AI-based systems \cite{nastjuk_what_2020,llorca2021trustworthy}, including the ability to audit and analyze them to understand why certain decisions have been taken \cite{Dwivedi2023a}, that is, to introduce a form of model transparency.
There are two main approaches towards this transparency: interpretable models and black-box explanations.
Interpretable models are white-box and their decision rules can be directly interpreted, such as tree-based models like random forests.
Black-box models, including all forms of neural networks, do not allow direct inspections, instead they are explained through additional functionality and techniques from explainable AI (XAI).
Interpretable models have the advantage of not needing external tooling for transparency, which adds overhead and is potentially misleading \cite{Rudin2022}. At the same time, performance-wise neural networks and their variants are the state-of-the-art in many domains.

A challenge inherent to both classes of models is that there can be many well-performing models trained from the same dataset, yet with large differences in their internal structures and encoded rules, nevertheless producing the same predictions.
This effect is called the \textit{Rashomon effect} in machine learning \cite{breiman_statistical_2001}.
It does not only affect model structure and model complexity \cite{fisher_all_2019}, but it has been shown to affect model explanations, too.%

Closest to our work, \citet{DBLP:conf/pkdd/MullerTBJBW23} were the first to highlight the significance of the Rashomon effect for explainable machine learning, demonstrating its effect on several case studies.
They train models to give exact same outputs on the dataset, while starting from different initializations, and can still show significant disagreement in the model explanations.
Although conducted in a controlled setting, their work highlights the potential pitfalls of the Rashomon effect for the reliability of XAI.

In this work, we investigate the existence of the Rashomon effect for explainable scene understanding in the automated driving domain, trained on a real-world dataset.
Starting from symbolic scene graphs in the form of Qualitative Explainable Graphs (QXG) \cite{belmecheri2024trustworthy}, we train multiple models that provide good explanations for actions in scene graphs and analyze the alignment of explanations among the models, comparing them to human-labeled scene information.
We consider both interpretable random forest models \cite{belmecheri2024trustworthy} and black-box graph neural networks (GNN) \cite{qxgnn2025} as prediction models, covering both perspectives on the transparency challenge.
Experiments are performed using the nuScenes dataset \cite{Caesar2020} with additional labels from DriveLM \cite{sima2023drivelm}.

In summary, we make the following contributions:
(1) we show the existence of the Rashomon effect in scene understanding,
(2) we analyze explanation agreement within two distinct modeling approaches—one interpretable and one black-box—that use a qualitative scene representation to identify relevant objects,
and (3) we contextualize the observed results for trustworthiness and model inspection, highlighting opportunities for future research.

\section{Background}

This section presents the necessary background for our study.
It gives an overview and formalization of the Rashomon effect in general and it's definition for the context of our work.
Additionally, it describes the qualitative explainable graph (QXG), the graph-oriented scene representation used as the data format for the ML models in our experiments.

\subsection{Rashomon Effect}
Introduced and coined by Leo Breiman \cite{breiman_statistical_2001}, the \emph{Rashomon effect} refers to a multiplicity of high-performing but \emph{different} models for the same prediction task, i.e., the same train-validation-test split of data. The effect captures the multiplicity of valid \emph{interpretations} of the data, which Breiman highlighted in the varying attribution scores of the input features in his original work.  

A \emph{Rashomon set} is a set of models that have been trained using the same training data, with an approximately equally high general performance. Definitions of the Rashomon set are typically tailored to the purpose of the study, resulting in varying definitions.\citet{semenova_existence_2022} derive mathematical properties from the multiplicity of models able to interpolate a training set. Their definition of the Rashomon set is the set of model architectures capable of fitting the training set within some error, while traditional definitions encompass the set of models performing well on unseen data during training. As highlighted by Breiman, the Rashomon effect has consequences for the \emph{reliance} of feature importance attributions. \citet{fisher_all_2019} propose a model class reliance measure that quantifies the variation in feature importance given a model class. In this work, they present the following definition of the Rashomon set: 

\begin{definition}[Population $\epsilon$-Rashomon Set]\label{def:population-rashomon-set}
    Given a general data point $(x, y)\sim p_\mathcal{D}$, where $x$ is an input drawn from some input space $\mathcal{X}$, $y$ is an output drawn from some output space $\mathcal{Y}$ and $p_\mathcal{D}$ is the true joint population distribution, the Rashomon set $\mathcal{R}$ is a function of the Rashomon parameter $\epsilon > 0$, a loss function $L\qty(f(x), y)\to\mathbb{R}$, a reference model $\tilde{f}$, and a model space $\mathcal{F}$, defined as 
    \begin{equation}
        \mathcal{R}\qty(\epsilon) = \qty{f\in\mathcal{F}: \mathbb{E}\qty[L\qty(f(x),y)]\leq\mathbb{E}\qty[L\qty(\tilde{f}(x), y)] + \epsilon}, 
    \end{equation}
    where the expectation is taken over the population distribution. 
\end{definition}

Here, $\epsilon$ adjusts the maximum difference in loss to account for a model considered to be equally performing, and therefore included in the Rashomon set. 
The problem with \ref{def:population-rashomon-set} is the intractability of computing the expectation over the true population distribution, which is unknown. 
A more practical definition will therefore apply a test set, unseen to the model during training, to give an empirical estimate of the population loss. 

In this work, we work with the following definition: 
\begin{definition}[Validation $\epsilon$-Rashomon Set]\label{def:validation-rashomon-set} 
    Given a training set, $\mathcal{D}_{\text{train}}=\qty{x_j, y_j}_{j=1}^M$, and a validation set, $\mathcal{D}_{\text{val}} = \qty{x_i, y_i}_{i=1}^N$, the Rashomon set $\mathcal{R}$ is a function of the Rashomon parameter $\epsilon > 0$, a loss functional $L\qty(f)\to\mathbb{R}$, a reference model $\tilde{f}$, and a model space $\mathcal{F}$, defined as 
    \begin{equation}
        \mathcal{R}\qty(\epsilon) = \qty{f\in\mathcal{F}: \hat{\mathbb{E}}_{\mathcal{D}_{\text{val}}}\qty[L(f)]<\hat{\mathbb{E}}_{\mathcal{D}_{\text{val}}}\qty[L(\tilde{f})] + \epsilon}, %
    \end{equation}
    where $\hat{\mathbb{E}}_{\mathcal{D}_{\text{val}}}\qty[L(f)]$ denotes the average loss over the validation set, and all models are trained using the same training set, $D_{\text{train}}$. 
\end{definition}

\subsection{Qualitative Explainable Graphs}
The Qualitative Explainable Graph (QXG) serves as a scene representation, capturing the qualitative spatial-temporal relationships among objects in a graph structure \cite{belmecheri2023acquiring,belmecheri2024trustworthy}. 
It has applications to symbolic cause determination \cite{fvl2025} and data-driven relevant object identification with graph neural networks \cite{qxgnn2025}.
For a given scene, denoted as $s$, the corresponding graph is constructed with a unique node for each object within the set of detected objects in the scene, $\mathcal{O}$. 
Edges, contained in the set $\mathcal{V}$, are established between any two objects that co-appear in at least one frame $f$.

\begin{figure}[t]
    \centering
    \includegraphics[width=\columnwidth]{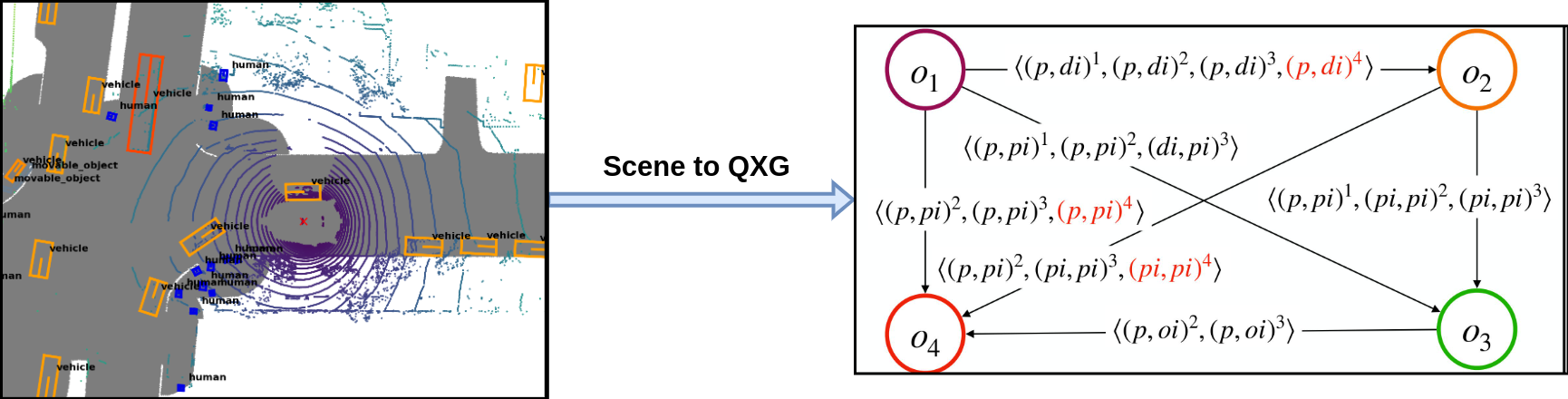}
    \caption{Illustration of the QXG built from a scene.}
    \label{fig:qxgbuilder}
\end{figure}

The labels on these edges are derived from spatial relations that are evaluated for each object pair $(o_i,o_j) \in \mathcal{O}$ at every frame. 
Formally, this is achieved using a set of algebras $\mathcal{A}=\{A_1,\ldots, A_n\}$. 
Each algebra $A_i$ is defined by a set of relations $R_i=\{r_1,\ldots,r_m\}$, with each relation $r_i$ describing a particular spatial property between objects $(o_i,o_j) \in \mathcal{O}$ at a specific frame $f_k$. 
Since a scene $s_i \in S$ comprises multiple frames, a new relation $r_k$ is determined from each algebra $A_i$ for every frame $f_k$, which endows the QXG with spatio-temporal properties. 
Figure~\ref{fig:qxgbuilder} provides an illustration of a QXG built from a three-frame scene; while only the third frame is highlighted, the QXG itself contains information on the dynamics of the spatio-temporal interactions across all three frames.

In this paper, we employ a combination of three qualitative calculi to model the relationships between objects and capture the necessary spatial information. 
These are: \textbf{Qualitative Distance Calculi} \cite{renz_qualitative_2007} for distance, \textbf{Qualitative Trajectory Calculi} \cite{dylla_survey_2015} for trajectory dynamics, and \textbf{Rectangle Algebra} \cite{renz_qualitative_2007} for relative positioning. 

Consider the following example:

\begin{example}
In a given frame $f_k$, consider two detected objects, $o_1$ (a car) and $o_2$ (a pedestrian), with their respective bounding boxes $bbox_1$ and $bbox_2$. 
If we apply the \textbf{Qualitative Distance Calculi}, the set of possible relations could be $\textbf{\{very close, close, far, very\ far\}}$. 
The specific relation is determined by computing the Euclidean distance between the centroids of $bbox_1$ and $bbox_2$. 
The relation is then set according to whether the value of $distance(bbox_1,bbox_2)$ falls below, exceeds, or lies between thresholds $(\theta_1, \theta_2, \theta_3)$.
\end{example}

However, it is crucial to note that the QXG's underlying formulation and its application are not dependent on this specific choice of calculi. 
The framework remains valid as long as the selected calculi possess sufficient expressiveness to describe, at a minimum, the relative positioning and distance of objects. 
For certain use cases, the representation can be further enriched by integrating additional calculi \cite{dylla_survey_2015}.

\section{Method}

In this section, we describe the action explanation problem in scene understanding and describe two approaches for action explanation using qualitative explainable graphs as the scene representation.

\subsection{Problem Description}

\begin{definition}[Action Explanation]
    Given a scene graph $G = (O, V) $, and an observed action $a$ performed by one of the objects in the scene, identify the object $o \in O$ that has caused the action.
\end{definition}
Put differently, the action explanation problem consists of identifying the relevant object as the cause for an action. 
The problem is a simplification of the real world in such that actions can be multi-causal or stem from unobserved circumstances, i.e., events within a vehicle or irrational personal decisions.
Nevertheless, it captures the essence of the complexity and is suitable for our study, especially since there are no extensive datasets with multi-cause labels that would support studying an extended problem setting.

For the purpose of our study, we consider two action explanation settings that approach the problem from different technical perspectives. The first decomposes the scene graph and uses more interpretable rule-based models, whereas the second works on the full graph but uses more complex graph neural networks.

\subsection{Pair-based Action Explanation}\label{sec:rf_action_explanation}

\begin{figure}[t]
    \centering
    \includegraphics[width=0.6\columnwidth]{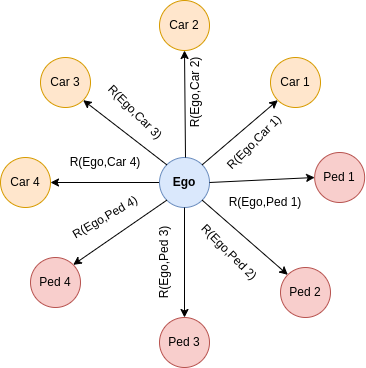}
    \caption{Extracted star graph centered around action-performing object.}
    \label{fig:stargraph}
\end{figure}

\begin{figure}[t]
    \centering
    \includegraphics[width=0.9\columnwidth]{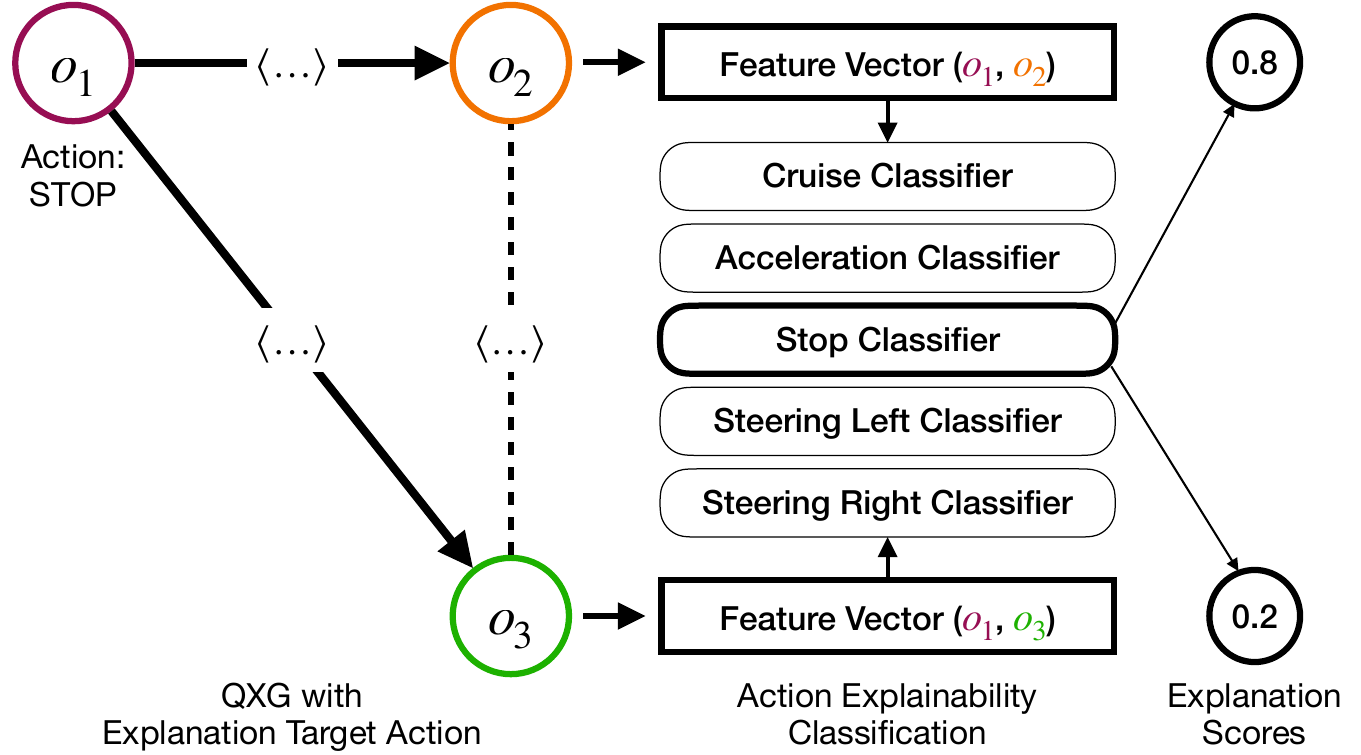}
    \caption{Edge-list-based Action Explanation (adapted from \citet{belmecheri2024trustworthy})}
    \label{fig:rf_action_explanation}
\end{figure}

We follow the indirect approach from \citet{belmecheri2024trustworthy} and approach the action explanation problem as a multi-classification problem on all object pairs, consisting of the object having performed the action, e.g., the ego vehicle, and every other object in the scene graph (see Figure~\ref{fig:rf_action_explanation}).
This approach first extracts from the overall scene graph a star graph centered around the action-performing object (see Figure~\ref{fig:stargraph}), which then gets deconstructed into an edge list of all object pairs.
Each object pair is classified according to which action it might have caused by a rule-based model, like a decision tree, a random forest, or gradient-boosting decision trees, as in our case.
The most relevant object is the object with the highest classification score for the actually observed action.

Decomposing the graph into an edge list has the advantage that each edge list is of fixed size and can be easily handled by common ML models, like the random forests used in our setting.
However, it comes at the cost that there is no shared knowledge about the present objects in the scene, as each object is classified individually and only the explainability scores are aggregated then. 
Due to the varying number of objects in each scene, this is not trivially realized for models with a fixed input size.
To overcome this limitation, other models can be applied, such as the graph neural network discussed in the next section.

Due to the interpretability of the classification model, the rules and reasoning for the selection of the most relevant object having caused the action are accessible, and the overall decision process is transparent to inspection. 
However, given the size of practical and performant tree-based models, they may require in-depth analysis and alternative explanation techniques, as they are used for full black-box models, to become sufficiently transparent.

\subsection{Graph-based Action Explanation}

\begin{figure}[t]
    \centering
    \resizebox{\columnwidth}{!}{\begin{tikzpicture}[
    box/.style={rectangle, draw, minimum width=2.5cm, minimum height=1cm, align=center, rounded corners},
    embedding/.style={box, fill=pink!20},
    attention/.style={box, fill=blue!20},
    extraction/.style={box, fill=green!20},
    decision/.style={diamond, draw, fill=green!20, align=center, minimum width=3cm, minimum height=2cm},
    arrow/.style={-{Stealth}, thick},
    node distance=1.5cm
]
    \node[embedding] (node_feat) {Node Features};
    \node[embedding] (edge_feat) [below=0.75cm of node_feat] {Edge Features};
    
    \node[embedding] (node_emb) [right=of node_feat] {Node Embedding};
    \node[embedding] (edge_emb) [right=of edge_feat] {Edge Embedding};
    
    \node[attention] (gat1) [right=3cm of node_emb] {GAT Layer 1};
    \node[box] (relu1) [right=of gat1] {ReLU};
    
    \node[attention] (gat2) [right=of relu1] {GAT Layer 2};
    \node[box] (relu2) [right=of gat2] {ReLU};

    \node[decision] (star) [below=4.5cm of node_feat] {Star Graph\\Extraction};
    \node[extraction] (concat) [right=of star] 
        {Concatenate Relations\\$[x_0, e_{0j}, x_j]$};
    
    \node[box] (hidden) [right=of concat] {Hidden Linear};
    \node[box] (relu3) [right=of hidden] {ReLU};
    \node[box] (output) [right=of relu3] {Output Layer};
    \node[box] (class) [right=of output] {Action`\\Classification};

    \draw[arrow] (node_feat) -- (node_emb);
    \draw[arrow] (edge_feat) -- (edge_emb);
    \draw[arrow] (node_emb) -- (gat1);
    \draw[arrow] (edge_emb) -- (gat1);
    \draw[arrow] (gat1) -- (relu1);
    \draw[arrow] (relu1) -- (gat2);
    \draw[arrow] (gat2) -- (relu2);

    \draw[arrow] (relu2.south) -- ++(0,-2.8) -| (star);

    \draw[arrow] (star) -- (concat);
    \draw[arrow] (concat) -- (hidden);
    \draw[arrow] (hidden) -- (relu3);
    \draw[arrow] (relu3) -- (output);
    \draw[arrow] (output) -- (class);

    \begin{scope}[on background layer]
        \node[fit=(node_feat)(edge_feat)(node_emb)(edge_emb),
              draw, dashed, inner sep=0.5cm] (emb_box) {};
        \node[above=0.2cm of emb_box] {Feature Embedding};
        
        \node[fit=(gat1)(relu1)(gat2)(relu2),
              draw, dashed, inner sep=0.5cm] (gat_box) {};
        \node[above=0.2cm of gat_box] {Graph Attention Layers};

        \node[fit=(star)(concat),
              draw, dashed, inner sep=0.5cm] (rel_box) {};
        \node[above=0.2cm of rel_box] {Relation Extraction};
        
        \node[fit=(hidden)(relu3)(output)(class),
              draw, dashed, inner sep=0.5cm] (class_box) {};
        \node[above=0.2cm of class_box] {Classification};
    \end{scope}
\end{tikzpicture}}
    \caption{Graph Neural Network architecture for action classification (adapted from \citet{qxgnn2025}).}
    \label{fig:gnn}
\end{figure}
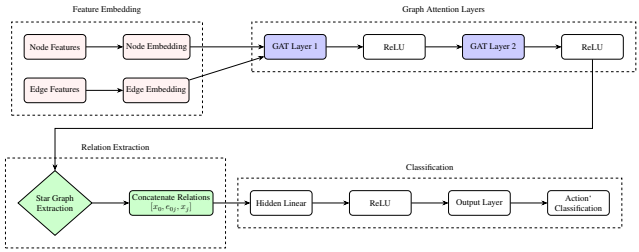

We mimic the interpretable action explanation approach, but introduce additional context on the objects in the scene through a graph neural network as the classification model.
Specifically, we modify the GNN architecture previously introduced for action-independent relevant object identification \cite{qxgnn2025} (see Figure~\ref{fig:gnn}).
Whereas the previous architecture performed binary classification of each object's relevance, our modified architecture is trained on the observed actions and we extract the relevant action-causing objects through explainable ML techniques.

The process works by classifying the scene according to which action is expected to occur.
Then the feature attribution according to the true, observed action -- which might differ from the predicted action -- is calculated to identify the node with the highest attribution, that is, the object deemed most relevant for the observed action.
Ways to calculate feature attribution include integrated gradients \cite{DBLP:conf/icml/SundararajanTY17} (used in the experiments), Shapley values-based techniques like SHAP \cite{DBLP:conf/nips/LundbergL17}, or other primary attribution technique.

Using the GNN broadens the context of the prediction by incorporating information from each object in the scene, but it comes at the cost of a non-interpretable model and the need for external feature attribution.
In the following experiments, we will evaluate whether there is a difference in the ambiguity of the provided explanations within and among the two classes of models.

\section{Experimental Setup}

\subsection{Experiments}

We execute two experimental settings, following the two approaches for action explanation, (1) \textit{pair-based action explanation} and (2) \textit{graph-based action explanation}.
For each approach, we collect a multiplicity of good models, the validation Rashomon set. We then analyze the existence and prevalence of the Rashomon effect by how well their explanations align with each other.

\subsection{Dataset}
For the experimental validation, we rely on the nuScenes dataset \cite{Caesar2020} with relevance annotations from DriveLM \cite{sima2023drivelm}. This combination was selected as it is the only available resource that provides labels for object relevance in driving scenarios.

The nuScenes dataset is a large-scale collection of real-world autonomous driving videos, captured across diverse geographic locations and weather conditions. Its comprehensive annotations include 3D bounding boxes for objects such as vehicles, pedestrians, and cyclists in every frame.

Complementing this, the DriveLM dataset from OpenDriveLab is specifically designed to understand driver behavior and object importance. It enhances a subset of scenarios from nuScenes by labeling objects based on their relevance to a driver's decision-making.

In nuScenes, each video is 20 seconds long and the DriveLM annotations can annotate multiple relevant objects at different times in the video.
We generate QXGs of 5 frames for all videos in nuScenes for which there is an actual action of the ego vehicle. The set of actions we consider is the combination of \{going straight, steering to the left, steering to the right\} $\times$ \{driving fast, driving normal, driving slow, not moving\}. 
These actions could be handled separately, but we choose the joint action space to simplify the setup for the experiments and the models.
In total, this setup results in 2131 scenes with ego vehicle actions and annotated relevant objects.

\subsection{Model Selection}

We aim to find a validation $\epsilon$-Rashomon Set for the dataset for the interpretable and the black-box action explanation settings.
Since both explanation settings are different, we address their training and the selection of the Rashomon set separately rather than as a single population of models.

For the interpretable action explanation, we train 100 gradient boosting decision trees \cite{friedman2001greedy} as the classifier with stochastic feature selection and subset sampling (bagging) \cite{friedman2002stochastic}, which avoids overfitting, but introduces randomness into the training process. 
To select the Rashomon set, we set $\epsilon=0.05$, meaning we include all models with at least 95\% performance of the best model's performance on the validation set.
As the models are very similar in their performance (within 2\% performance), this leads to all 100 models being well-performing and being part of the Rashomon set.

For the black-box case, we train 116 GNNs, too, for 200 epochs each.
Again, we set $\epsilon=0.05$, which leads to a selection of 32 well-performing models.

For all models, the train/validation split is fixed, but the randomness in model training, e.g., for data splits and model initialization, is seeded differently per model training.

\subsection{Data Collection}

After the selection of well-performing models, the next step is to collect explanation and measure the models' agreement.
We run all selected models on the scenes with human-annotated relevance labels to explain the action.
From the object identified as relevant, we extract the feature importance of the spatio-temporal features encoded in the QXG using SHAP and rank them by their absolute importance.
That means, for each of the scenes with an annotated relevant object we run the prediction models, check their correctness, and collect their explanations through an assessment of the feature contributions from SHAP \cite{DBLP:conf/nips/LundbergL17}.

We quantify explanation agreement using two metrics: Fleiss' kappa and Kendall's W (described below). 
We then report aggregated statistics (mean, standard deviation, min, max) for these metrics across all test scenes.

\subsection{Metrics}

The goal of the study is to understand the agreement in perceived feature importance among the models in the Rashomon set.

\subsubsection{Fleiss' Kappa}

Fleiss' kappa is a metric to measure the inter-rater agreement among multiple categorizations of items, i.e., how much multiple raters agree in their classification \cite{fleiss1971measuring}. 
In our case, we measure how much multiple models agree in their ranking of the most important features for a prediction, i.e. we categorize the features as relevant or not.
Fleiss' kappa is then expressed as the agreement of the top-rated features over a random selection:
\begin{equation}
\kappa = \frac{P_o - P_e}{1 - P_e}
\end{equation}

In this formula, $P_o$ is the observed proportional agreement among the models, representing the mean agreement calculated across all features. The term $P_e$ is the expected agreement that would occur if the models made their classifications purely by chance, based on the overall distribution of ``relevant'' and ``not relevant'' classifications, according to a uniform distribution \cite{randolph2005free}.

The resulting value of kappa indicates the level of agreement, where a value of 1 signifies perfect agreement among all models, a value of 0 suggests the agreement is no better than random chance, and a negative value indicates that the observed agreement is worse than chance, implying systematic disagreement. 
Generally, higher values of kappa indicate stronger agreement among the models on which features are most important.

For our experiments, we consider a feature "important" if it ranks among the top-k features. 
We calculate kappa for varying $k \in [1,20]$, i.e., the top half of features.

\subsubsection{Kendall's W}

Kendall's W, also known as Kendall's Coefficient of Concordance, is a non-parametric statistic used to assess the agreement among several raters \cite{kendall1962rank}. It is useful to determine the consensus in ordered data. In our context, it measures how consistently multiple models rank the features based on their importance, moving beyond simple categorization to evaluate the entire ordering.

Kendall's W is calculated based on the sum of squared deviations of the grand sum of ranks for each feature from the mean. The formula is \(\displaystyle W = \frac{12S}{m^2(n^3 - n)}\).

Here, $S$ represents the sum-of-squares statistic over the rank sums, $m$ is the number of raters (our models), and $n$ is the number of items being ranked (the features). Each feature is assigned a rank by each model, and $S$ is computed from the totals of these ranks for each feature.

The value of W ranges from 0 to 1, where a value of 1 indicates perfect agreement, meaning all models produced the exact same ranking of features. 
A value of 0 signifies a total lack of agreement, suggesting that the rankings provided by the models are essentially random with respect to one another. 
Unlike correlation coefficients that can be negative (such as Fleiss' kappa), W does not measure disagreement, only the presence or absence of agreement. 
Therefore, a higher value of W signifies a stronger consensus among the models regarding the relative importance of the features.

\subsection{Technical Setup}

Our experiments are implemented in Python, using LightGBM's gradient boosting decision trees \cite{DBLP:conf/nips/KeMFWCMYL17}, scikit-learn's utilities \cite{scikit-learn}, and PyTorch \cite{DBLP:conf/asplos/AnselYHGJVBBBBC24} and PyTorch-Geometric \cite{Fey/Lenssen/2019} for the GNN. Explanations are calculated via the Captum library \cite{kokhlikyan2020captum}.
Experiments are performed on a PC with an Intel Xeon E3-1246 CPU, 24\,GB RAM and a NVIDIA GeForce RTX 2060 GPU.\footnote{Online: \url{https://doi.org/10.5281/zenodo.17045085}}

\section{Results}

We evaluate the models on their predictions on the labeled part of the nuScenes dataset, as described above.
Specifically, we analyze their predictions in terms of the agreement on the most influential features, as estimated through SHAP.

\subsection{Top-K Most Influential Features}

In the first analysis, we compare the selection of the most influential, top-k features per scene among the models.
Here, we only consider if the same features are part of the top-k most influential features, without considering their order.

\begin{figure*}[t]
    \centering
    \includegraphics[width=0.95\textwidth]{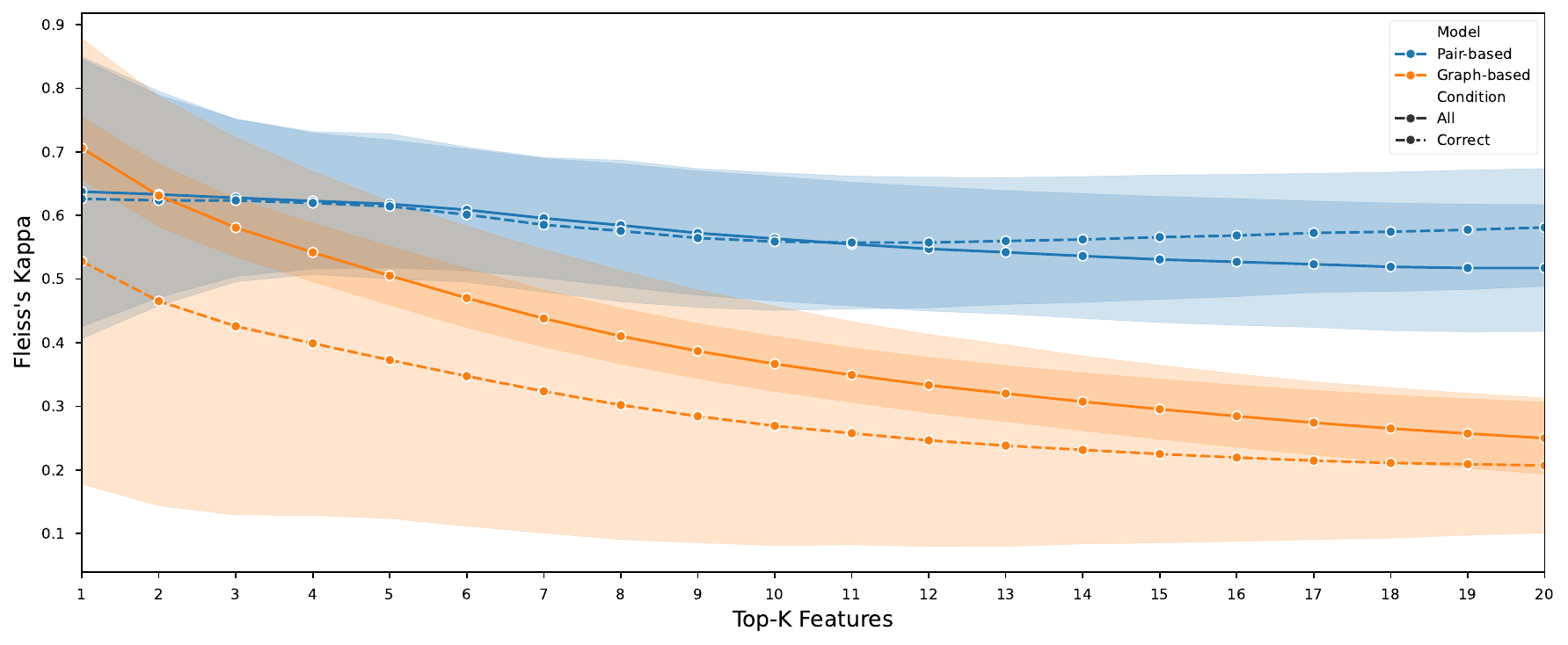}
    \caption{Agreement on top-k features among models. We observe a higher agreement for the Pair-based LightGBM than the Graph-based GNN.}
    \label{fig:kappa}
\end{figure*}

We perform the analysis for all $k \in [1,20]$, which covers the top half of features, and calculate Fleiss' kappa, i.e., the agreement among the models on which features belong into the top-k. There are 42 features in the pair-based approach and 45 features in the graph-based approach, where the difference stems from a variation in the preprocessing of the rectangle algebra, which gets decomposed into x- and y- positioning in the graph-based approach, and the information on the object types in the pair-based approach, which is part of the node information in the graph-based approach.

The results are shown in Figure~\ref{fig:kappa}.
Generally, the agreement in the pair-based approach is much higher than in the graph-based approach, except for the first feature.
After that, the agreement in the graph-based approach declines, whereas the pair-based approach is steadier.
This means that there is more variation in the lower-rank features for the graph-based approach with more fluctuation around the top-k-border, whereas the most influential features are relatively steady among the models with a kappa of 0.6--0.9.
This is a reassuring result in that there is an agreement on which features influence the decision the most on what is the relevant object that has caused an action.
However, the more features we consider in the ranking, the more disagreement we observe, which is potentially misleading for further analysis of the models, possibly pointing to spurious correlations that are only important for one model, but not for the problem in general.
This could indicate that there are \textit{symmetries in the data}, i.e., equivalently relevant features to take a decision if a human were asked, although it is unclear if they are true alternatives or if there are spurious correlations in the training set that are registered by some models.

To better understand the results, we separate the results between all predictions made for an action, including those that highlight the wrong object as relevant (respectively not the one that was labeled as relevant), and only those predictions that correctly identify the labeled relevant object.
Interestingly, we observe a difference between the two approaches here.
In the pair-based approach, the agreement is very similar, independent of the correctness, with a gap for higher top-k values towards a stronger agreement among correct predictions.
Wrong predictions might indicate a weakness of that specific model for the given scenario, raising doubts about the reliability of the explanations, whereas correct predictions seem to have a better general understanding of the scene.

Surprisingly, for the GNNs, the agreement among explanations for correct predictions is consistently lower than the agreement across all predictions, albeit with a much larger variance (the shaded area).
Again, we might see some symmetries in the data.
Otherwise, where the tree-based models of the pair-based approach are constructed according to the training dataset (stopping the depth of additional boosting once the performance plateaus), the capacity of a neural network is set upfront, not entirely independent of the training dataset (as part of the hyperparameter selection), but not coupled to the training process itself.
This could be caused by an \textit{overparameterization}, where the model has many redundant pathways to represent the same function, resulting in similar predictions from different internal reasoning.

\subsection{Feature Ranking Agreement}

In the second analysis, we compare the agreement on the ranking of all features.
In each analysis, we present the results for both experimental settings jointly, while discussing and highlighting their specifics individually.

\begin{table}[t]
    \centering
    \adjustbox{max width=\columnwidth}{%
\begin{tabular}{llrrrr}
\toprule
Model & Condition & Mean & Std. Dev. & Min & Max \\
\midrule
Pair-based & All & 0.32 & 0.09 & 0.08 & 0.61 \\
Pair-based & Correct & 0.40 & 0.07 & 0.24 & 0.61 \\\midrule
Graph-based & All & 0.07 & 0.02 & 0.02 & 0.24 \\
Graph-based & Correct & 0.14 & 0.18 & 0.03 & 1.00 \\
\bottomrule
\end{tabular}}
    \caption{Kendall's W for the feature ranking, separated per explanation approach and prediction correctness. Even though the models are all similarly well-performing, their agreement on feature importance is low (graph-based) to moderate (pair-based) only.}
    \label{tab:kendallw}
\end{table}

To provide a more holistic view of explanation agreement, we use Kendall's W to measure the consensus on the entire feature ranking, not just the top-k features. The results are summarized in Table~\ref{tab:kendallw}.
The pair-based models show a moderate level of agreement, which improves noticeably when considering only correct predictions. This suggests that when these models correctly identify the relevant object, their underlying reasoning is more consistent.

In contrast, the GNNs exhibit very low agreement on the feature ranking. 
While the agreement doubles for correct predictions, it remains low overall. The high standard deviation and maximum value of 1.0 for correct GNN predictions indicate that in some specific cases, the GNNs can agree perfectly, but on average, their internal reasoning for the same correct prediction is highly divergent. 
This reinforces the finding from the top-k analysis: different GNNs in the Rashomon set might learn different, yet equally effective, predictive functions.

\subsection{Discussion \& Limitations}

In this section, we interpret our observed results and draw some conclusions on how to embrace or mitigate them.
As part of the result discussion, two potential causes for model agreement were raised -- \textit{Symmetries in the data} and \textit{Overparameterization} --, which we will discuss specifically.
To put the results into context, we discuss the limitations of our experiments and outline paths for further research to advance and broaden them.

\paragraph{Symmetries in the Data}

An agreement of the models would be easy to interpret, as it confirms with our expectations.
Same circumstances should lead to the same conclusions for the same reasons.
Disagreements are more difficult. Even if the models draw the correct conclusions, are they doing it for other -- equally correct -- reasons or for the wrong reasons, i.e., because they were badly trained?
For example, there can be multiple causes in a scene to stop the car, albeit the person crossing the crosswalk or the ball rolling on the street.
Our experiments cannot answer this questions as we are lacking the necessary data in our dataset.
Future work could overcome this by acquiring more data or performing additional studies on varied data, e.g., leaving out objects previously identified as relevant, to understand the relevance of prominent features under data variations.

\paragraph{Overparameterization}

On the one hand, overparameterization is a generalization of the symmetry issue.
A model might cover many ways to explain a prediction, some if which might be correct according to the provided, while there is not sufficient data to exclude the wrong ones.
On the other hand, it is a technical problem as the model under training is not suited for the task at hand.
It might be too small, relying always on the same few features and not grasping nuances from others.
It might be too big, learning too many variations, but not their constraints as there is no limitation, which could potentially be explained by being a form of shortcut learning \citep{geirhos_shortcut_2020}.
Where tree-based model training has its capacity bundled to its performance when using early stopping, this is difficult for black-box models like the used GNNs and it would require a dedicate hyperparameter selection procedure, which is potentially difficult.
For the case of automated driving and under consideration of nuScenes and DriveLM, an alternative could be to enforce explanation plausibility into the training paradigm, making a shift from performance optimization, i.e., achieving highest classification scores, towards plausible predictions, i.e., confirming to human expectations. In the case of overparameterization, this is an additional constraint on the final network, reducing the amount of technical symmetries and variations.
At the same time, it puts more dependency on the plausibility and reliability of the human annotations -- which is a problem itself, especially when considering decisions in automated driving.

\paragraph{Limitations}
Our study is biased by the focus on the Validation Rashomon set (rather than the focus on the training performance), the choice of $\epsilon$ (smaller = tighter and maybe more comparable), that we consider practical models, even though they do not achieve a perfect score (unlike \citet{DBLP:conf/pkdd/MullerTBJBW23}, which explore the extreme point of models with perfect score on their training sets).
This approach is more oriented towards a real-world application of the models, where the main selection criterion is their validation score.
Potentially, from our definition, we would expect a lower model disagreement than when overfitting the models on the training data, which is an additional study that could be performed as a conformity check of our results.
Finally, our analysis is constrained to one type of interpretable model (gradient boosting) and one GNN architecture. 
The extent of the Rashomon effect may vary with different model families.

\section{Conclusion}

In this paper, we empirically quantified the impact of the Rashomon effect on the stability of explanations for action prediction in autonomous driving. Using a symbolic Qualitative Explainable Graph (QXG) representation, we trained and analyzed Rashomon sets of two distinct model classes: interpretable, pair-based gradient boosting models and complex, graph-based GNNs.

Our findings serve as a cautionary tale against the naive acceptance of post-hoc explanations. 
To achieve true trustworthiness in AI systems, we must move beyond the illusion of a single "ground-truth" rationale for each prediction.
The path forward requires a paradigm shift: instead of asking ``Why did this model make this prediction?'', we should ask ``What are the possible reasons a good model could have for this prediction?''. 

Future work should not seek to eliminate this multiplicity, but rather to understand and leverage it. 
Promising research directions include: 
1) deepening our analysis of the effect, e.g., by addressing inherent explanation sensitivity \citep{yeh_fidelity_2019}; 
2) developing techniques to find a ``consensus explanation'' that captures the features most consistently identified as important across the Rashomon set; 
3) using explanation variance as a new form of uncertainty quantification; and 
4) designing training regimes that regularize models to produce more stable explanations, e.g., by using human relevance annotations as an auxiliary loss.
By being aware of and embracing the multiplicity of rationales, we can develop a more robust and trustworthy approach to interpreting AI.

\section{Acknowledgments}
This work is funded by the AutoCSP project of the Research Council of Norway (grant number 324674) and the European Commission through the projects: 
AI4CCAM (Trustworthy AI for Connected, Cooperative Automated Mobility, grant agreement No 101076911), 
MARS (Manufacturing Architecture for Resilience and Sustainability, grant agreement No 101091783), 
and AI4COPSEC (Boosting EU Copernicus Security and Maritime
Monitoring with AI and Machine Learning (ML), grant agreement No 101190021).

\bibliography{refs}

\end{document}